\documentclass[journal]{IEEEtran}

\usepackage{xcolor,soul,framed} 

\colorlet{shadecolor}{yellow}
\usepackage[pdftex]{graphicx}
\graphicspath{{../pdf/}{../jpeg/}}
\DeclareGraphicsExtensions{.pdf,.jpeg,.png}

\usepackage[cmex10]{amsmath}
\usepackage{array}
\usepackage{mdwmath}
\usepackage{mdwtab}
\usepackage{eqparbox}
\usepackage{url}

\begin{document}
    \title{Systematic Evaluation of Hip Exoskeleton Assistance Parameters for Enhancing Gait Stability During Ground Slip Perturbations}
  \author{Maria T.~Tagliaferri and
      Inseung~Kang,~\IEEEmembership{Member,~IEEE}\\
      
  \thanks{M. T. Tagliaferri and I. Kang are with the Department of Mechanical Engineering, Carnegie Mellon University, Pittsburgh, PA, 15213 USA (e-mail: {\tt\footnotesize mtagliaf@cmu.edu}).}
  }


\maketitle
\begin{abstract}
\textit{Objective:} Falls are the leading cause of injury-related hospitalization and mortality among older adults. Consequently, mitigating age-related declines in gait stability and reducing fall risk during walking is a critical goal for assistive devices. Lower-limb exoskeletons have the potential to support users in maintaining stability during walking. However, most exoskeleton controllers are optimized to reduce the energetic cost of walking rather than to improve stability. While some studies report stability benefits with assistance, the effects of specific parameters, such as assistance magnitude and duration, remain unexplored. \textit{Methods:} To address this gap, we systematically modulated the magnitude and duration of torque provided by a bilateral hip exoskeleton during slip perturbations in eight healthy adults, quantifying stability using whole-body angular momentum (WBAM). \textit{Results:} WBAM responses were governed by a significant interaction between assistance magnitude and duration, with duration determining whether exoskeleton assistance was stabilizing or destabilizing relative to not wearing the exoskeleton device. Compared to an existing energy-optimized controller, experimentally identified stability-optimal parameters reduced WBAM range by 25.7\% on average. Notably, substantial inter-subject variability was observed in the parameter combinations that minimized WBAM during perturbations. \textit{Conclusion:} We found that optimizing exoskeleton assistance for energetic outcomes alone is insufficient for improving reactive stability during gait perturbations. Stability-focused exoskeleton control should prioritize temporal assistance parameters and include user-specific personalization. \textit{Significance:} This study represents an important step toward personalized, stability-focused exoskeleton control, with direct implications for improving stability and reducing fall risk in older adults.

\end{abstract}


\IEEEpeerreviewmaketitle

\section{Introduction}

\IEEEPARstart {D}{uring} daily locomotion, the human body experiences occasional disturbances in gait stability, such as walking on a slippery surface \cite{bruijn2013locomotion_stability}. Maintaining stability during these perturbations requires rapid coordination between sensory feedback and motor responses to perform stabilizing gait adjustments and prevent falls \cite{JKPT2021,VANDENBOGERT2002199}. With aging, declines in muscle strength and cognitive function can reduce both the speed and effectiveness of these corrective responses \cite{10.1093/ageing/afp200,hardwick2022age}. Consequently, falls during walking are a major source of injury in older adults, with nearly one in four individuals over the age of 65 requiring hospitalization each year due to fall-related incidents \cite{woollacott1997balance}. 

To address this, substantial research over the past decade has explored developing and evaluating strategies to improve walking stability and reduce fall risks \cite{Chang680,Sun2018}. In addition to balance-specific training, mobility aids, such as canes and walkers, are commonly prescribed to individuals for safe navigation across daily environments. However, these devices cannot actively respond to destabilizing events and require continuous user attention and proper placement to be effective, which may be compromised during unexpected perturbations \cite{BATENI200474,Atoyebi2019}. These limitations highlight the need for assistive technologies that can actively augment user stability and respond to perturbations without relying on voluntary user actions. Wearable lower-limb robotic exoskeletons are a promising solution \cite{10479575}. Unlike passive walking aids, exoskeletons can detect user instability in real time and deliver stability-enhancing torque assistance to lower-limb joints \cite{10964407,11063070,Monaco2017}. Hip exoskeletons are particularly promising, because assistance at the hip can directly influence step width and length, as well as trunk kinematics, which are key contributors to stability recovery during perturbations.

Despite their potential to improve gait stability and reduce fall risk, existing exoskeleton literature has predominantly focused on reducing the energetic cost of walking \cite{7487663,10858452,Martini2019}. This is mainly due to difficulties in directly quantifying gait stability and the high biomechanical variability associated with human response to perturbations. As a consensus continues to develop on optimal stability metrics for exoskeleton evaluation, some recent studies have begun to design exoskeleton controllers specifically for assisting gait stability \cite{10452751,Afschrift2023}. However, these studies have mainly evaluated single control strategies or fixed parameter sets. To our knowledge, no existing research has systematically examined how variations in assistance parameters such as assistance timing or magnitude affect stability measures \cite{Monaco2017}. Understanding these relationships is critical for developing exoskeleton control strategies that can effectively respond to destabilizing disturbances and improve stability outcomes.

To evaluate the effects of exoskeleton assistance parameters on gait stability, we selected whole-body angular momentum (WBAM) as the primary outcome measure. We chose WBAM because it has been widely used and validated in prior studies of human locomotion and perturbation responses \cite{leestma2023linking,10.1371/journal.pone.0230019}. WBAM serves as a holistic metric of dynamic balance by quantifying the summed angular momentum of all body segments relative to the center of mass (COM). Humans regulate WBAM within a narrow range during steady-state locomotion \cite{10.1242/jeb.008573}, therefore examining deviations in WBAM during perturbations can provide critical insights into stability control. In populations at high risk of falls, increases in WBAM range have been correlated with reduced performance on clinically relevant balance metrics and decreased walking stability \cite{NOTT2014129,PIJNAPPELS2005388}. Furthermore, prior work has demonstrated that minimizing WBAM range during slip events may reduce the destabilizing effect of perturbations \cite{10.1371/journal.pone.0230019}. We evaluated sagittal plane WBAM, given that both the perturbation and exoskeleton assistance are designed to act in this plane.

In this study, we investigate how varying the time duration and magnitude of exoskeleton assistance influences biomechanical responses during ground slip perturbations. Slip perturbations were chosen because they are among the most common types of perturbations, and a leading cause of falls among older adults \cite{Shimada2003}. This work will explore two main hypotheses. (1) WBAM range will decrease with assistance magnitude, with larger magnitudes yielding greater stability benefits. This relationship reflects the idea that increasing torque magnitude provides greater resistance to perturbation forces, resulting in smaller changes in lower-limb kinematics \cite{8601369}. (2) The optimal duration of assistance will align with the biological flexion response to the slip. We expect WBAM will decrease as duration approaches this timing and increase again at longer durations. This is supported by previous findings that balance interventions should reinforce, but not override, the user’s own recovery strategies \cite{beck2023exoskeletons,Afschrift2023}. Our sub-hypothesis is that the best-performing assistance parameters will outperform a previously validated baseline controller designed for improving energetics during steady-state walking. Given that most current exoskeleton controllers are optimized for energetic outcomes during steady-state gait, we examine how this class of control influences WBAM during perturbation responses. We expect the baseline controller to be less effective, as human control strategies during steady-state walking differ substantially from those employed during unstable conditions \cite{MCANDREW2011644,YOO201998}.

\section{Robotic Hip Exoskeleton}
\label{sec:Robotic Exoskeleton}
\subsection{Mechatronic Design}
\begin{figure}[t!]
    \centering
    \includegraphics[width=1\linewidth]{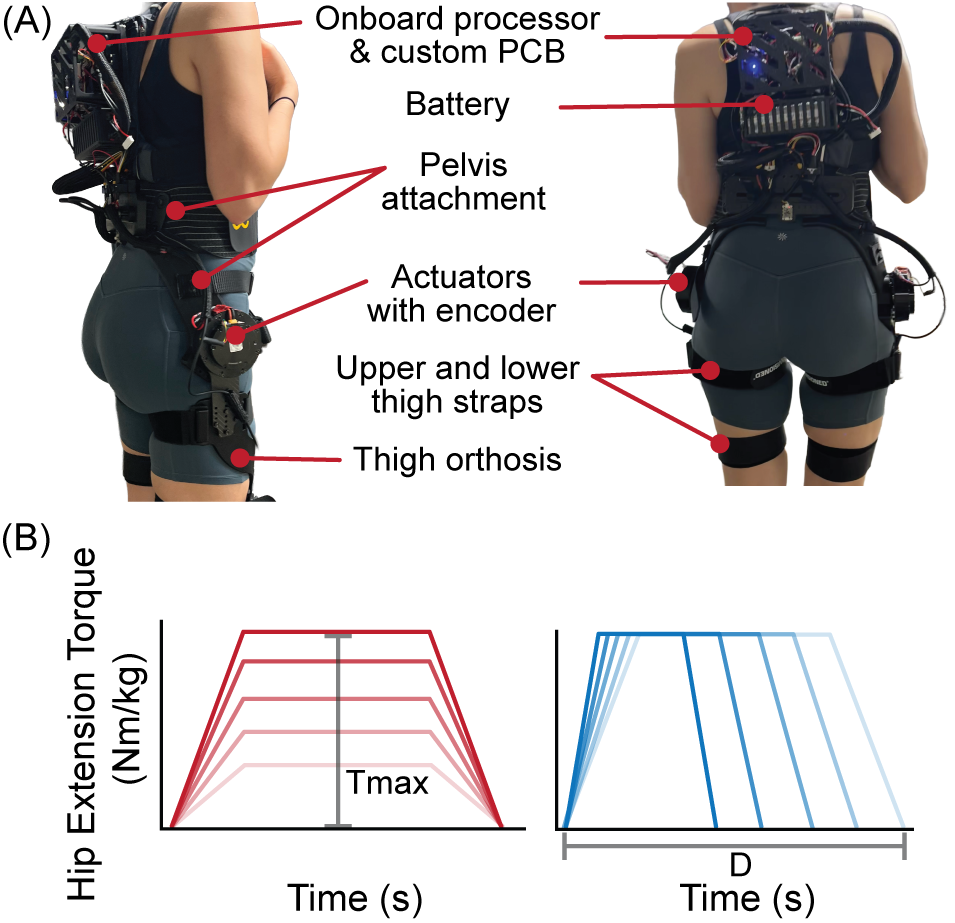}
    \caption{Robotic hip exoskeleton designed to assist the user’s hip flexion and extension during locomotion. (A) Actuators located at each hip joint are controlled by an onboard microprocessor. Adjustable orthotic shells and pelvic strapping secure the device to the user. (B) A trapezoidal hip extension torque profile was applied at the hip. Five magnitude and five duration values were chosen to evaluate the assistance parameter space. Informed consent for publication of this image was obtained from the participant.} 
    \label{fig:fig1}
\end{figure}

We developed a bilateral robotic hip exoskeleton to deliver sagittal-plane torque assistance at the hip joint during treadmill-simulated slip perturbations (Fig. \ref{fig:fig1}A). The electromechanical system incorporates two brushless DC motors with built-in encoders (AK80-9, CubeMars), a 22.2 V Lithium-Polymer battery, and a 3D printed nylon-based user interface. Control algorithms were executed on an onboard microprocessor (Jetson Orin Nano Devkit, Nvidia), supported by a custom printed circuit board that facilitated communication with the sensors and actuators. The full system weighs 4.5 kg.

The exoskeleton provides 100$^\circ$ of hip flexion and 30$^\circ$ of hip extension in the sagittal plane, with an additional 5$^\circ$ of passive hip abduction and adduction enabled by a flexible carbon-fiber joint. At both hip joints, the actuators can produce up to 18 Nm of torque. The device attaches to the user at four locations: the upper and lower thigh via an orthotic shell, the pelvis band, and the electronics backpack. Each attachment point is adjustable to accommodate a range of body sizes.

\subsection{Control Framework}
Following a slip, reactive hip joint moments primarily act to slow the sliding motion of the foot and limit the anteroposterior displacement of the trunk. Previous studies have shown that applying torque at the hip can effectively aid this recovery process \cite{Monaco2017,Manzoori2024}. However, the timing and magnitude of torque assistance have not been systematically explored to understand and optimize the stabilizing effect. To systematically evaluate the impact of these control parameters, we selected a time-dependent trapezoidal control profile (Fig. \ref{fig:fig1}B). Torque assistance was applied in the extension direction to counteract the increased hip flexion caused by our forward slip perturbation. The profile was defined as,

{\small
\begin{equation}
\tau(x) =
\begin{cases}
T_{\max}\left(\dfrac{x}{0.2D}\right), & 0 \le x \leq 0.2D, \\
T_{\max}, & 0.2D < x < 0.8D, \\
T_{\max}\left(1 - \dfrac{x - 0.8D}{0.2D}\right), & 0.8D \le x \le D.
\end{cases}
\end{equation}
}

where \(x\) is time (\(x=0\) ten milliseconds after onset of perturbation), \(T_{\max}\) is the torque scaling factor and \(D \) is the duration of assistance. A trapezoidal profile was selected because it enables independent control over assistance magnitude and duration while avoiding the abrupt discontinuity exhibited in square-wave inputs, which can introduce actuator instability.

The assistance onset was set to 10 ms following perturbation onset to account for system communication latency. Previous studies have shown that keeping the delay below the user’s physiological balance response time, approximately 120 ms on average for healthy adults \cite{7281223}, improves the effectiveness of exoskeleton-mediated stability assistance \cite{beck2023exoskeletons}.

The baseline assistance profile was an eight-parameter spline developed by Franks \textit{et al}., originally designed to minimize metabolic cost through real-time optimization using experimental data \cite{franks2021comparing}. The control parameters defined key features of the torque profile, including onset timing, peak timing, peak magnitude, and the curvature of the rising and falling phases. The peak magnitude was set to 20\% peak biological torque based on prior exoskeleton research demonstrating both the effectiveness and broad applicability of this assistance level \cite{doi:10.1126/scirobotics.adi8852}. Gait phase was estimated in real time using ground reaction forces \cite{ROERDINK20082628}, and the spline profile was defined based on the average stride length of the first five strides preceding the perturbation. This control profile was selected as the baseline because it has been widely applied in hip exoskeleton control across a range of locomotor tasks \cite{Kim2022,Bryan2021}.

\section{Methods}
\label{sec:Methods}
\subsection{Experimental Protocol}
\begin{figure*}[!t]
    \centering
    \includegraphics[width=1\textwidth]{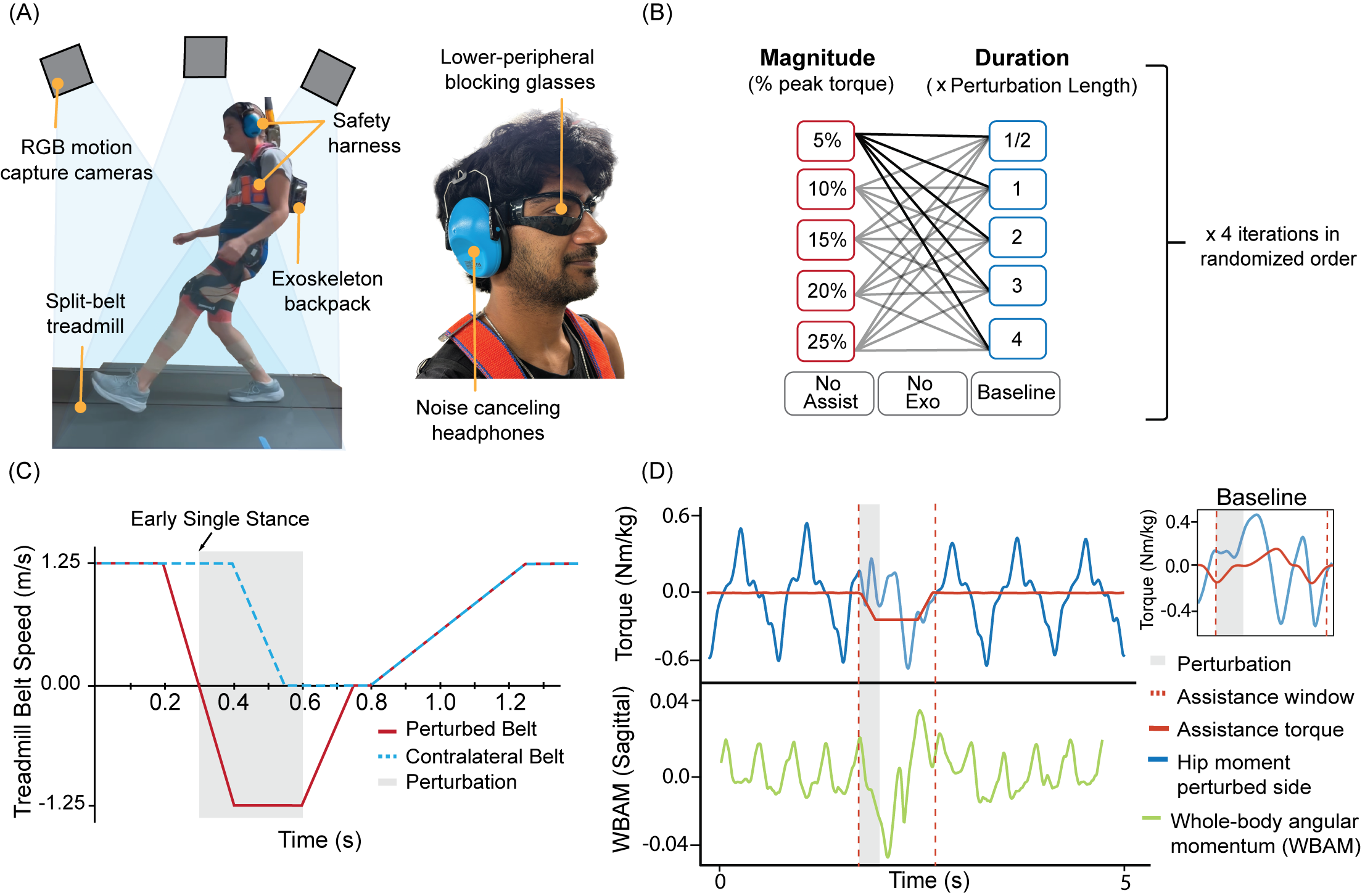}
    \caption{Experimental setup, protocol, and representative data. (A) Subjects experienced anteroposterior slip perturbations during treadmill walking while wearing the robotic exoskeleton. Anticipatory cues were minimized using noise-canceling headphones and visual-obscuring glasses. (B) The experimental design comprised 28 different assistance conditions, repeated across four sessions. (C) Treadmill belt velocity profile used to elicit the slip response. (D) Time-series data from a representative trial illustrating biological hip torque (blue), exoskeleton assistance (red), and the resulting WBAM (green), relative to the perturbation duration (gray area). Informed consent for publication of this image was obtained from the participants.}
    \label{fig:fig2}
\end{figure*}

Eight able-bodied adult participants (3 female and 5 male; age, 22.75$\pm$2.64 years; height, 174.8$\pm$6.9 cm; body mass, 66.4$\pm$9.8 kg) were recruited for this study. All participants provided informed consent prior to participation and the study protocol (STUDY2024\_00000506) was approved by the Carnegie Mellon University Institutional Review Board.

Each subject wore a robotic hip exoskeleton (Fig. \ref{fig:fig1}A) while walking at 1.25 m/s on a force-instrumented split-belt treadmill (Bertec, USA) and experiencing unexpected anteroposterior slip-like perturbations (Fig. \ref{fig:fig2}A). The exoskeleton was commanded to provide torque assistance only during the perturbation event to avoid interfering with baseline steady-state WBAM.

Assistance was applied to the perturbed leg under one of 28 experimental conditions (Fig. \ref{fig:fig2}B), which included two torque profile types: a trapezoidal profile (Fig. \ref{fig:fig1}B) and a Piecewise Cubic Hermite Interpolating Polynomial spline profile \cite{franks2021comparing}. For the trapezoidal profile, five duration parameters were selected: 0.5, 1, 2, 3, and 4 times the perturbation length. These values were chosen to be centered around the peak timing of the biological flexion response to the slip perturbation, which pilot data showed to occur at approximately twice the perturbation length ($\sim$600 ms). Five assistance levels were selected: 5\%, 10\%, 15\%, 20\%, and 25\% of the peak biological hip extension moment, normalized by subject body weight, based on previous exoskeleton literature \cite{Molinaro2024}. A “no assistance” condition, in which the device was worn but with no torque applied, was also tested. In addition, a “no exoskeleton” condition served as a control condition by exposing participants to the slip perturbation without the device.

The perturbation was designed to simulate a sudden slip and was delivered during the early single-stance phase of the perturbed leg (Fig. \ref{fig:fig2}C). The perturbation was achieved by an abrupt, transient change in the velocity and direction of the belt corresponding to the perturbed limb \cite{7582950}. This method replicated the destabilizing dynamics of real-world slip events while allowing precise timing control and repeatability across trials \cite{siragy2023comparison},\cite{Lee2025}. To ensure safety, participants wore an overhead harness (Fig. \ref{fig:fig2}A), which was height-adjusted to prevent interference during steady-state walking. Once the perturbation began, a trigger was sent to the exoskeleton to provide torque assistance (Fig. \ref{fig:fig2}D). 

Each 15-second walking trial consisted of a single perturbation (Fig. \ref{fig:fig2}D), paired with one of the 28 assistance conditions. The conditions were presented in a randomized order and repeated four times per subject, with a 10-minute seated rest between each session to minimize fatigue. This protocol resulted in a total of 112 perturbation trials per subject (Fig. \ref{fig:fig2}B). Perturbation onset timing and limb (left or right leg) were pseudo-randomized and controlled using a custom software program that interfaced with both the treadmill (Fig. \ref{fig:fig2}C) and the motion capture system (Lock Lab, Vicon, UK). To further minimize anticipatory reactions, subjects wore noise-canceling headphones to block auditory cues from belt movements and specialized glasses that obscured the lower peripheral view of the treadmill belts (Fig. \ref{fig:fig2}A). The experiment began with a 10-minute adaptation phase during which subjects practiced walking with the exoskeleton and glasses at the target speed. Subjects were also given a single practice perturbation trial to reduce potential bias associated with the first experimental condition.

Trials were recorded using markerless motion capture cameras (Vicon, UK) for post-hoc processing of WBAM. Additionally, we quantified user perception of the various assistance conditions using the modified Orthotics and Prosthetics Users’ Survey (OPUS) scoring method \cite{Heinemann2003, Jarl2012}. At the end of each trial, subjects were asked to rate their perceived stability during the perturbation on a 5-point scale, where 1 indicated “very unstable” and 5 indicated “very stable.”

\subsection{Data Processing and Analysis}
Raw data were first pre-processed to reduce signal noise and remove motion artifacts. Ground reaction force signals were filtered using a fourth-order Butterworth filter (6 Hz cutoff) \cite{doi:10.1126/scirobotics.adi8852}.

\subsubsection{Segmented Body Model}
A full-body biomechanical model was generated using markerless motion capture software (Theia3D, Theia Markerless Inc.), comprising 16 body segments and 54 internal degrees of freedom. The model included the trunk, pelvis, upper arms, forearms, hands, thighs, shanks, and feet. All joints were modeled as spherical, with joint centers defined according to established anatomical conventions. For each segment $i$, a local orthogonal reference frame was placed at its COM, and the proximal and distal end positions were estimated based on the biomechanical model. Segmental rotational matrices, as well as linear and angular velocities, were calculated by importing the 16-segment model into a motion capture data analysis tool (Visual3D) and performing inverse kinematics. All kinematic data were filtered using the same procedure applied to the ground reaction forces.

Body segment inertial parameters (mass, COM location, and mass moment of inertia) were estimated following the literature standard \cite{Hanavan1964}. The exoskeleton mass was incorporated into the model by distributing it evenly between the thigh segments and the torso. Segmental mass and COM positions were then used to estimate the whole-body COM. For each subject and trial, data were partitioned to capture activity before, during, and after perturbation onset for analysis. Trials were excluded if a jumping response occurred, and gait events were manually labeled for trials exhibiting cross-stepping.

\subsubsection{Quantifying Gait Stability}
To investigate the effectiveness of each assistance strategy, we calculated WBAM range across the perturbed and recovery strides \cite{leestma2023linking}, \cite{CHAUMEIL2024112018}. The sagittal-plane WBAM was computed as:

\begin{equation}
WBAM =
\frac{
\sum_{i=1}^{16} 
\left[ 
\left( \overrightarrow{x_{i}} - \overrightarrow{x_{}} \right) 
\times m_i \left( \overrightarrow{v_{i}} - \overrightarrow{v_{}} \right) 
+ I_i \ast \overrightarrow{\omega_i}
\right]
}{\mathit{mvh}}
\end{equation}

From the 16-segment body model described above we extracted inertial properties ( \(I_i \) ), and the position ( \(\overrightarrow{x_{com}}\)\ ), linear velocity ( \(\overrightarrow{v_{com}}\)\ ), and angular velocity ( \(\overrightarrow{\omega_i}\)\ ) vectors of the COM of each segment in the sagittal plane. The denominator term normalizes the WBAM for cross trial comparison, where $m$ is the subject mass, $v$ is the average walking speed, and $h$ is the subject height. WBAM range was then calculated by subtracting the minimum from the maximum normalized WBAM value across the perturbed and recovery strides. Computing the range across both strides helped reduce bias arising from variability in gait cycle length following the perturbation, and provides an integrated rather than instantaneous measure. For each trial, WBAM range was normalized by the average WBAM range during the unperturbed gait cycles to account for trials with naturally higher baseline WBAM. For each assistance condition, the final WBAM range was obtained by averaging values across the four repeated trials. WBAM range is reported as a percentage change from the steady-state baseline, with 0\% representing steady-state WBAM. Consistent with existing literature, reductions in WBAM range were interpreted as improvements in stability.

\subsubsection{Interpolation of Discrete Parameters}
Given the involvement of human subjects, exhaustive sampling of the full parameter space for assistance magnitude and duration was not feasible. Accordingly, a limited set of discrete parameter combinations was tested. To estimate how assistance parameters influenced WBAM regulation at a finer resolution than directly sampled, we used interpolation to create a continuous 2-D response surface across subjects. Interpolation was performed using a Radial Basis Function (RBF) \cite{552059}, with a smoothing factor of 0.4 and epsilon of 0.1. The interpolation weights were inversely proportional to the trial-wise variances to account for both within- and between-subject variability. RBF interpolation was chosen over grid-based or polynomial interpolation because it preserves smoothness without imposing a predefined functional form, which is consistent with the continuous, nonlinear nature of WBAM responses to perturbations. The minimum of the interpolated surface, referred to as the theoretical optimal, was calculated to estimate the optimal parameter set. A contour map of the interpolated surface was generated to visualize the response. This procedure was applied in the same manner to the OPUS score data.

\subsection{Statistical Analysis}
A one-way repeated-measures analysis of variance (ANOVA) was conducted on the mean WBAM range across repetitions of the full parameter set to assess whether motor learning significantly influenced WBAM. 

The influence of varying assistance parameters on WBAM range was evaluated using a linear mixed-effects regression model. Assistance magnitude and assistance duration were modeled as fixed effects, with an interaction term included to evaluate whether their effects on WBAM were interdependent. Subject identity was included as a random intercept to account for repeated measurements and inter-subject variability. Models were fit using restricted maximum likelihood estimation. Parameter-specific linear regressions were subsequently performed to further characterize the interaction between assistance magnitude and duration.

To evaluate differences in WBAM range across assistance conditions, with a specific focus on comparison to the best-performing parameter set for each subject, a one-way repeated-measures ANOVA was performed. The four conditions tested were: no-exoskeleton, no-assistance, baseline controller, and the best-performing parameter set. Post-hoc pairwise comparisons were conducted using Bonferroni correction. The no-exoskeleton and no-assistance conditions were included to assess the effect of wearing the exoskeleton.

To improve stability and quantify uncertainty of the interpolated response surfaces, 1,000 bootstrap surfaces were generated by resampling the experimental trials with replacement and re-fitting the RBF. The optimal assistance magnitude and duration were identified for each individual bootstrap surface, producing a bootstrap distribution from which 95\% confidence intervals were computed.

Residuals were tested for normality using Shapiro-Wilk tests to confirm that model assumptions were satisfied across all models.

\section{Results}
\label{sec:Results}
\subsection{Effect of Protocol Repetition}
\begin{figure}[!ht]
    \centering
    \includegraphics[width=1\linewidth]{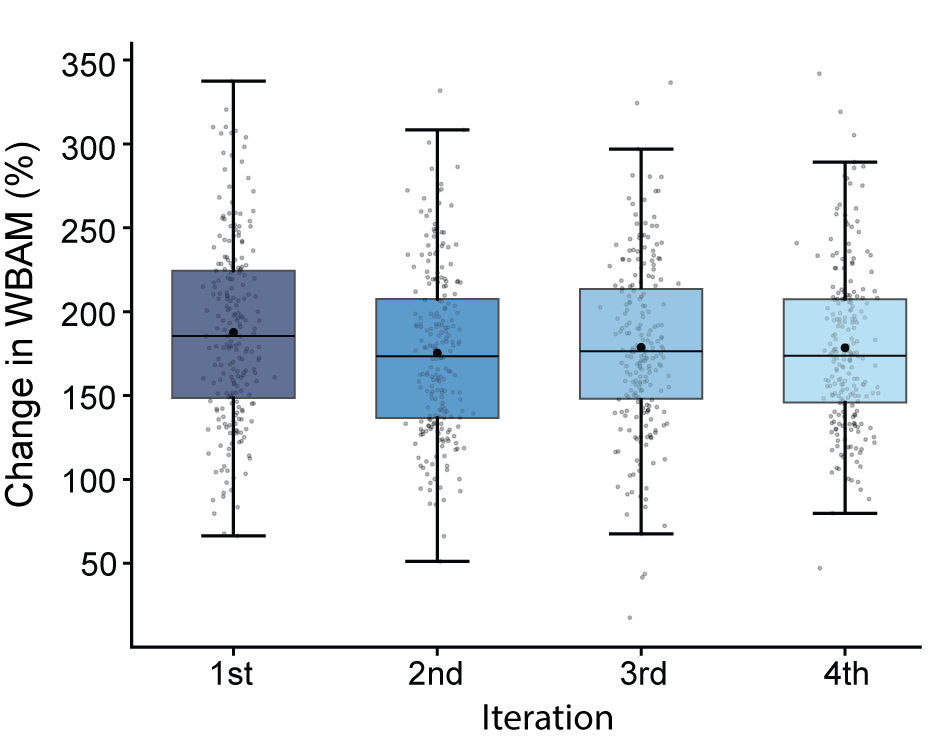}
    \caption{WBAM across protocol repetitions. The repeated-measures ANOVA indicated that no significant effect of session repetition on the percent change in WBAM range. Black dots indicate the the group mean while gray dots represent individual subject data points.} 
    \label{fig:fig3}
\end{figure}

 A repeated‑measures ANOVA revealed no significant effect (F(3,21)=1.33, $p$=0.29) of protocol repetition on percent change in WBAM range (Fig. \ref{fig:fig3}). This indicates that motor learning across repetitions did not significantly influence WBAM responses.
\subsection{Effect of Assistance Duration and Magnitude}

\begin{figure}
    \centering
    \includegraphics[width=1\linewidth]{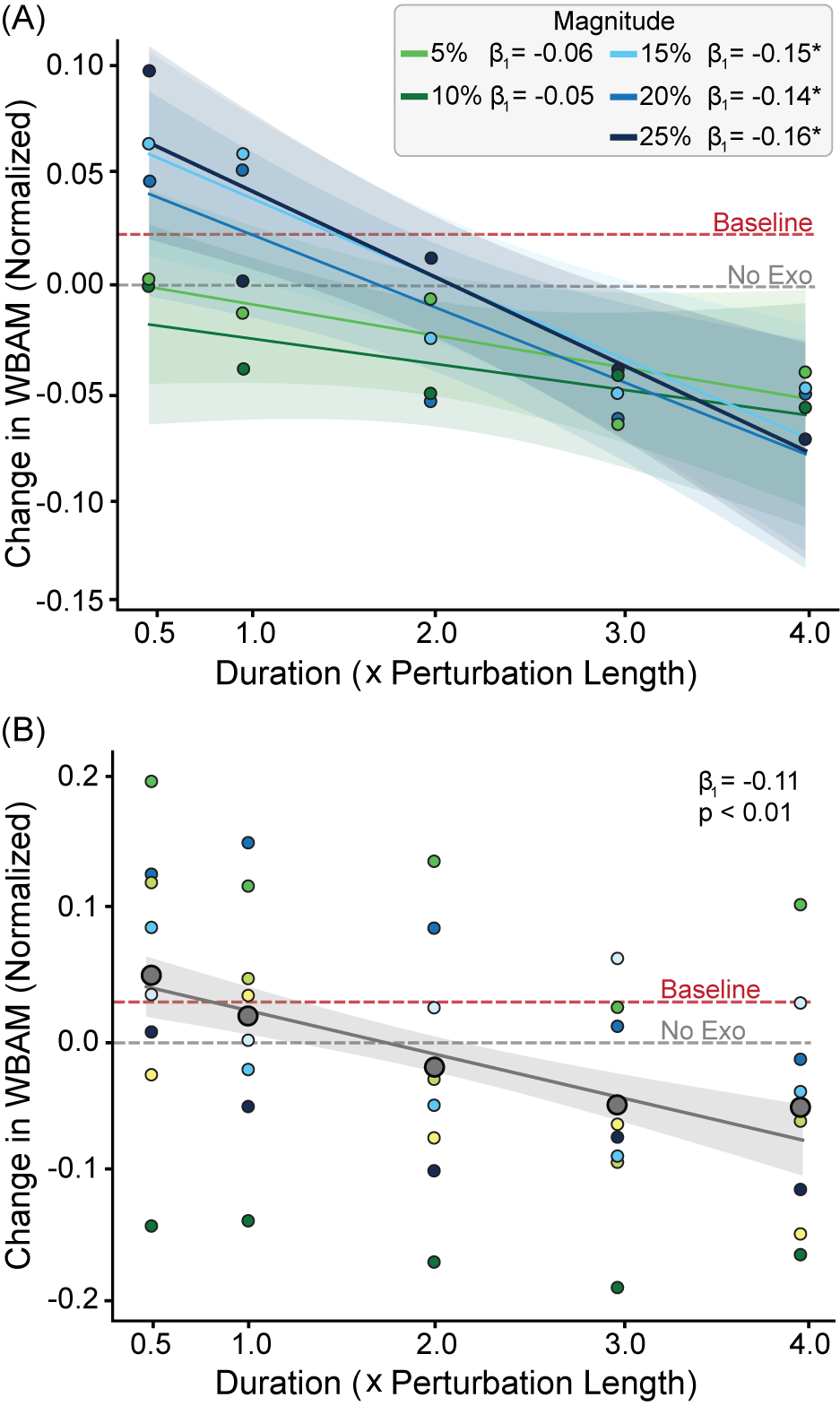}
    \caption{Influence of assistance duration and magnitude on gait stability. Decreases in WBAM range indicate improved stability. (A) Effect of assistance duration at fixed assistance magnitudes. (B) Effect of assistance duration aggregated across all magnitudes. Large gray dots represent group means and small dots indicate subject-specific averages. All data are normalized to no-exoskeleton condition. The dashed red line shows the mean WBAM range for the baseline controller condition, and shaded regions indicate 95\% confidence intervals. $\beta_{1}$ represents the slope of the line. Significance levels are indicated as: *$p$$\leq$0.01.} 
    \label{fig:fig4}
\end{figure}

The linear mixed‑effects model identified a significant effect of assistance magnitude on WBAM range ($\beta$=101.71, $p$=0.005) and a significant magnitude$\times$duration interaction ($\beta$=-104.30, $p$=0.032). Duration alone was not a significant predictor ($\beta$=-4.52, $p$=0.576). Substantial between-group differences in baseline WBAM range were observed (random-intercept variance = 1237.7), exceeding residual variance (350.1) and yielding a high intraclass correlation coefficient (ICC(1,1) = 0.78).

Parameter-specific linear regressions were used to evaluate the effect of assistance duration at fixed magnitudes (Fig.~\ref{fig:fig4}A). Data were normalized to each subject’s no-exoskeleton condition to illustrate performance relative to walking without the device. Significant negative linear trends for duration were observed at higher magnitudes (15, 20, 25\% peak torque; $p$=0.002, 0.004, 0.0003; $R^2$=0.06, 0.05, 0.08). At lower magnitudes (5\% and 10\%), slopes were small and nonsignificant ($p$$\geq$0.2).

Additionally, a linear regression of duration averaged across all magnitudes revealed a significant inverse relationship ($p$=0.00002) which is shown together with the subject-specific means to illustrate inter-subject variability (Fig.~\ref{fig:fig4}B). 

The linear regression of assistance magnitude at fixed durations exhibited weak, nonsignificant relationships with low coefficients of determination (all $R^2 $$\leq$0.02, $p$$\geq$0.05), except at the shortest duration ($\beta_{1}$=0.58, $p$=0.01). Therefore, these regressions were omitted from the figure.

For OPUS scores, the linear mixed‑effects model revealed a significant effect of assistance magnitude ($\beta$=-2.89, $p$=0.003) and a significant magnitude$\times$duration interaction ($\beta$=2.93, $p$=0.026). Duration alone was not a significant predictor ($\beta$=-0.21, $p$=0.325).

\subsection{Comparison of Assistance Conditions}
\begin{figure}
    \centering
    \includegraphics[width=1\linewidth]{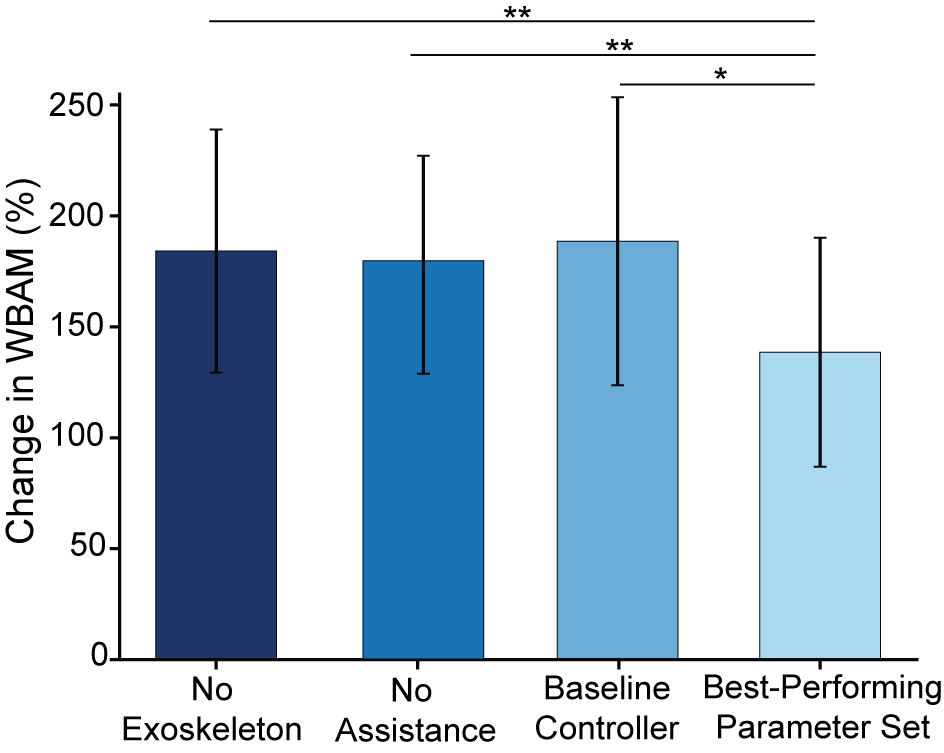}
    \caption{Average percent change in WBAM compared across conditions using a within-subject design. The best-performing parameter set represents the parameter set that produced the lowest percent change in WBAM range for each subject, averaged across subjects. Significance levels are denoted as: * $p$$\leq$0.05, ** $p$$\leq$0.01.} 
    \label{fig:fig5}
\end{figure}

The best-performing parameter set reduced WBAM range by 27.4±9.8\% relative to walking without the exoskeleton ($p$=0.005; Fig.~\ref{fig:fig5}) and by 25.7±11.4\% relative to the baseline controller ($p$=0.02). No significant differences were observed between the no-exoskeleton and baseline conditions.

The best-performing parameter set also significantly increased OPUS scores compared with the no-assistance, no-exoskeleton, and baseline conditions (all $p$$\leq$0.05).

\subsection{Interpolation of Discrete Parameters}
\begin{figure}[!ht]
    \centering
    \includegraphics[width=1\linewidth]{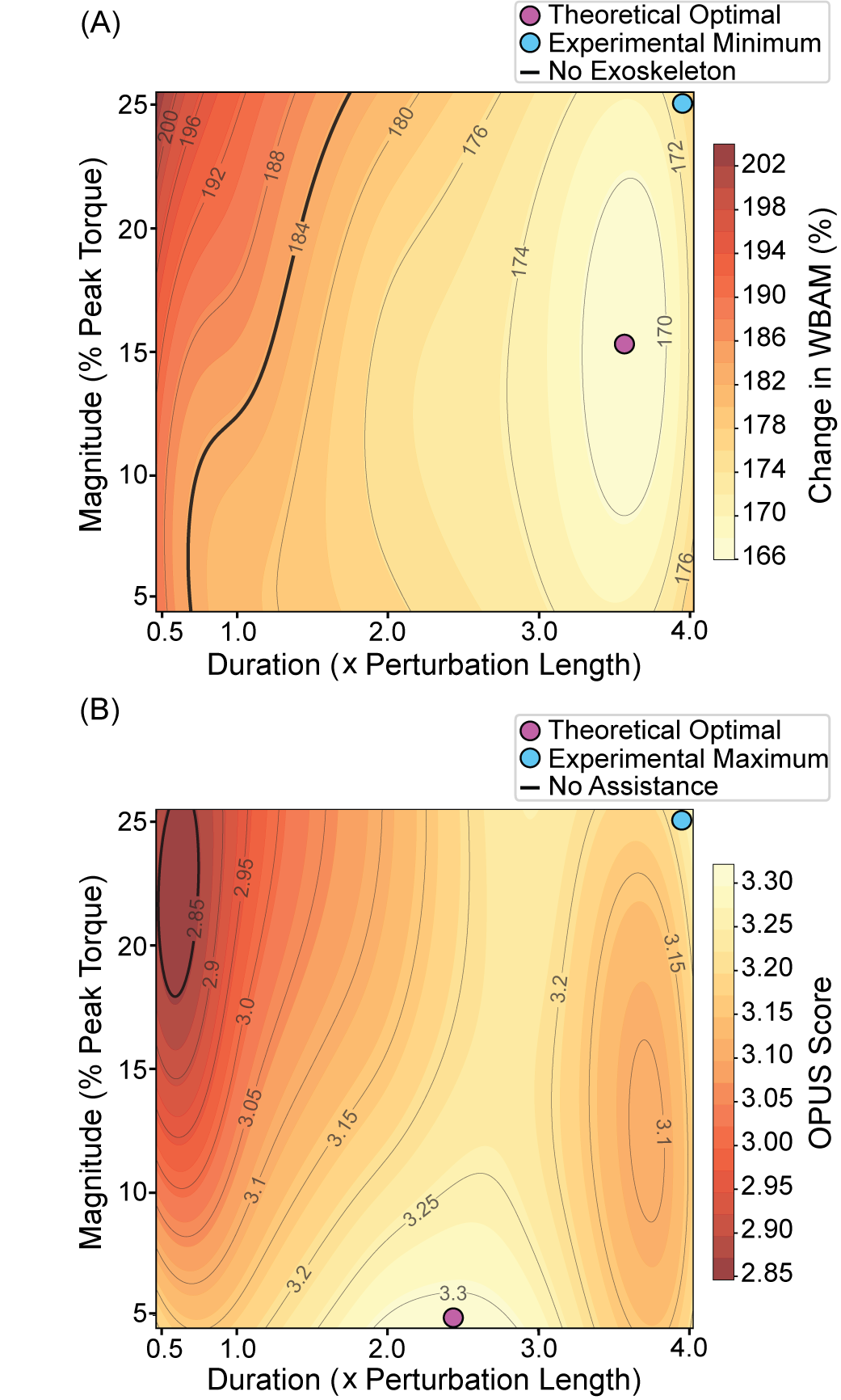}
    \caption{Interpolated response surfaces for objective and subjective stability. (A) Surface map of WBAM range across the assistance parameter space. The theoretical optimum (purple dot) represents the mean of the bootstrap-resampled surfaces, while the experimental minimum/maximum (blue dot) represents the best-performing discrete parameter set. (B) Surface map of perceived stability (OPUS score). Higher scores indicate greater user-perceived stability.} 
    \label{fig:fig6}
\end{figure}

The minimum of the interpolated response surface for WBAM range occurred at 15.9\% peak biological torque and 3.64 times the perturbation length (Fig.~\ref{fig:fig6}A). Bootstrap resampling confirmed the stability of this optimum, producing 95\% confidence intervals (CI) of 10.1-21.6\% for assistance magnitude and 3.48-3.72 for perturbation duration. The wider CI for magnitude aligns with the prior analyses showing that the effect of magnitude is attenuated at longer durations. For the experimentally sampled data, the smallest percent change in WBAM range occurred at an assistance magnitude of 25\% of peak biological torque and a duration of four times the perturbation length.

The maximum value of the interpolated response surface for OPUS score occurred at 2.5 times the perturbation length, and 4.9\% peak torque, however the CI spanned the entire surface, indicating high inter-subject variability. For the experimentally sampled data, the highest average score occurred at an assistance magnitude of 25\% of the peak biological torque and a duration of 4 times the perturbation length (Fig. \ref{fig:fig6}B).

\section{Discussion}
The goal of this study was to investigate how modulating the duration and magnitude of hip exoskeleton assistance influences WBAM range during anteroposterior ground slip perturbations. Our primary finding was that WBAM response during slip is governed by a significant interaction between assistance magnitude and duration. Specifically, the stabilizing or destabilizing effects of hip exoskeleton assistance depend critically on how torque magnitude is distributed in time. Our results also emphasize that conventional energy-optimized exoskeleton controllers are insufficient for reducing WBAM range during perturbations.

\subsection{Effect of Assistance Magnitude and Duration}
We hypothesized that WBAM range would decrease with 1) increasing assistance magnitude and 2) as assistance duration approached the length of the initial biological hip flexion response to the perturbation. Neither hypothesis was supported.

The influence of assistance magnitude on WBAM range was dependent on the duration of application. At short durations, WBAM range increased with increasing torque magnitude. In this region of the parameter space, excessive angular impulse was generated and overcompensated the slip-induced backward momentum, driving forward trunk rotation and destabilizing the user relative to the no-exoskeleton condition. However, as assistance duration increased, the effect of magnitude became non-significant. Extending the duration of torque application reduced the instantaneous angular impulse and appropriately counteracted changes in WBAM. In this region of the parameter space, stability was improved compared to the no-exoskeleton condition.

Contrary to our expectation, the best-performing assistance duration exceeded the reactive flexion response, aligning more closely with the average length of the entire perturbed gait cycle. We found that WBAM continues to fluctuate beyond the initial flexion response, as the increased limb momentum drives backward pitching of the COM. Extending the duration of assistance allows the applied torque to counteract momentum changes at both the joint and COM levels, resulting in a greater overall reduction in WBAM range. Although regression indicated a negative linear relationship between duration and WBAM range across magnitudes, this does not imply that an indefinite increase in duration is optimal. Assistance applied substantially beyond the natural recovery window would interfere with steady-state gait mechanics, ultimately increasing WBAM rather than further reducing it. This interpretation is supported by the interpolated surface map, which estimated an increase in WBAM range at the longest duration values.

Overall, our results highlight the critical role of assistance duration in determining whether torque assistance supports or disrupts recovery of stability. Optimizing assistance duration may lower peak torque requirements, lowering device weight and cost while maintaining effective gait stabilization. This aligns with prior work on energetic outcomes, which found that exoskeleton assistance magnitude should be optimized for maximum energetic savings rather than increased to the highest possible level \cite{8601369}, and with studies demonstrating the importance of other temporal control parameters, such as assistance onset timing, in stability outcomes \cite{beck2023exoskeletons}.

\subsection{Comparison to the Baseline Controller}
In agreement with our sub-hypothesis, the best-performing parameter set significantly reduced WBAM range compared to the baseline assistance profile. Post-hoc analysis indicated that the baseline controller’s poor performance was attributable to its timing: on average, the flexion peak timing of the baseline profile coincided with the slip-induced hip flexion response, amplifying backward angular momentum and increasing WBAM range in the opposite direction during the recovery step. 

One interesting finding was that the best-performing parameters varied significantly between subjects. This high inter-subject variability highlights the need for subject-specific personalization of exoskeleton assistance during perturbations. This finding aligns with prior studies that demonstrated the importance of tailoring assistance profiles on a per-user basis for optimal energetic benefits \cite{franks2021comparing,11112638}.

\subsection{Perception of Stability}
The best-performing parameters were consistent across WBAM range and OPUS scores. The regions of poorest performance similarly aligned across metrics. This indicates that participants were generally perceptive to how assistance influenced stability, particularly at the extremes of the parameter space. Subject-specific OPUS-optimal parameters did not always coincide with WBAM minima, however, suggesting limited sensitivity to finer parameter variations. High inter-subject variability in the parameters that maximized OPUS score further highlighting the importance of personalized assistance parameters. 

Overall, the best-performing parameter set significantly increased OPUS scores relative to no-exoskeleton and baseline controller conditions, indicating that appropriately tuned assistance may enhance perceived stability. This may be particularly beneficial for populations such as older adults, where perception of stability has been linked to performance in locomotor tasks \cite{10.2522/ptj.20080036}.

\subsection{Study Impact and Limitations}
This study provides a unique contribution to the field of lower-limb exoskeleton control. Whereas previous work has predominantly evaluated a single controller with heuristic parameters \cite{Monaco2017,Manzoori2024,Afschrift2023}, our approach systematically swept assistance duration and magnitude to map their combined effects on stability. To our knowledge, this is also the first study to benchmark stability-focused control against an energetics-optimized baseline, and the first to demonstrate a significant reduction in WBAM relative to both. The findings of this study establish a foundation for future development of control strategies specifically targeting stability outcomes.

This study has several limitations. First, we focused exclusively on anteroposterior slip perturbations, which may limit generalizability. Second, practical time constraints in human-subject experiments limited evaluation to a discrete set of assistance parameters, potentially constraining our understanding of WBAM responses. Future work will focus on developing adaptive control profiles for a broader range of perturbation types, leveraging biomechanical simulation to reduce the burden on human subjects. Finally, perturbation onset was assumed to be known for experimental consistency; integrating online perturbation detection in future studies will address this real-world limitation \cite{10964407,11063070}.

\section{Conclusion}
By systematically examining the magnitude and duration of exoskeleton assistance, this study demonstrates that gait stability during slip perturbations is a function of temporal torque distribution rather than magnitude alone. Our results highlight the inadequacy of energetics-optimized controllers for stability tasks and emphasizes the necessity of task-specific, personalized exoskeleton parameter tuning. These findings provide a framework for the next generation of adaptive wearable robots capable of providing effective, real-time fall prevention in complex environments.

\section*{Acknowledgment}
The authors would like to thank Nancy Sahagun for her contributions to data collection.


%




\ifCLASSOPTIONcaptionsoff
  \newpage
\fi





\bibliographystyle{IEEEtran}
\bibliography{IEEEabrv,Bibliography}

\end{document}